\documentclass[conference,a4paper]{IEEEtran}
\IEEEoverridecommandlockouts

\usepackage[hidelinks]{hyperref}
\usepackage[cmex10]{amsmath}
\usepackage{amssymb,amsfonts}
\usepackage{dblfloatfix}

\usepackage[ruled,vlined]{algorithm2e}
\usepackage{graphicx}
\graphicspath{{Figures/PDF/}{Figures/PNG/}}

\usepackage{booktabs}
\usepackage{siunitx}
\usepackage[numbers,compress]{natbib}
\usepackage{texnames}
\usepackage{bm,bbm}
\usepackage{orcidlink}
\usepackage{graphicx}
\usepackage{multirow}
\usepackage{tablefootnote}
\usepackage{adjustbox}
\usepackage{array} 
\usepackage{pifont}
\newcommand{\cmark}{\ding{51}} 
\newcommand{\xmark}{\ding{55}} 
\newcommand{\yes}{\cmark}
\newcommand{\no}{\xmark}

\begin{document}

\title{Enhanced LULC Segmentation via Lightweight Model Refinements on ALOS-2 SAR Data 
\thanks{This work supported by AIST policy-based budget project ``R\&D on Generative AI Foundation Models for the Physical Domain". The ALOS-2 original data are copy-righted by JAXA and provided under the JAXA-AIST agreement.}
}

\author{
	\IEEEauthorblockN{
		Ali Caglayan\orcidlink{0000-0002-3408-8659},
		Nevrez Imamoglu\orcidlink{0000-0002-2661-599X},
		Toru Kouyama\orcidlink{0000-0002-1060-3986}
	}
	\IEEEauthorblockA{
		\textit{National Institute of Advanced Industrial Science and Technology (AIST)}, Tokyo 135-0064, Japan\\
		\{ali.caglayan, nevrez.imamoglu, t.kouyama\}@aist.go.jp
	}
}

\maketitle
\begin{abstract}
This work focuses on national-scale land-use/land-cover (LULC) semantic segmentation using ALOS-2 single-polarization (HH) SAR data over Japan, together with a companion binary water detection task. Building on SAR-W-MixMAE self-supervised pretraining \cite{sar_w_mixmae}, we address common SAR dense-prediction failure modes, boundary over-smoothing, missed thin/slender structures, and rare-class degradation under long-tailed labels, without increasing pipeline complexity. We introduce three lightweight refinements: (i) injecting high-resolution features into multi-scale decoding, (ii) a progressive refine-up head that alternates convolutional refinement and stepwise upsampling, and (iii) an $\alpha$-scale factor that tempers class reweighting within a focal+dice objective. The resulting model yields consistent improvements on the Japan-wide ALOS-2 LULC benchmark, particularly for under-represented classes, and improves water detection across standard evaluation metrics.
\end{abstract}


\section{Introduction}
\label{sec:intro}
Foundation models are typically enabled by large-scale pretraining, with masked image modeling (MIM) being particularly effective for learning transferable representations from unlabeled data \cite{liu2023mixmae, he2022masked, hong2024spectralgpt, wang2022advancing}. This direction has substantially advanced remote sensing representation learning, especially for optical RGB imagery (e.g., geography-aware \cite{wang2022advancing} or large-scale pretraining \cite{ayush2021geography}) and multispectral sensors \cite{hong2024spectralgpt, noman2024rethinking, li2024s2mae} via MAE-style \cite{he2022masked} objectives. However, translating the same paradigm to synthetic aperture radar (SAR) is less straightforward because SAR follows different imaging physics and statistics than optical data, and downstream performance is often bottlenecked by dense prediction quality rather than classification-level semantics.

\begin{figure*}[t!]
	\centering
	\includegraphics[width = \textwidth]{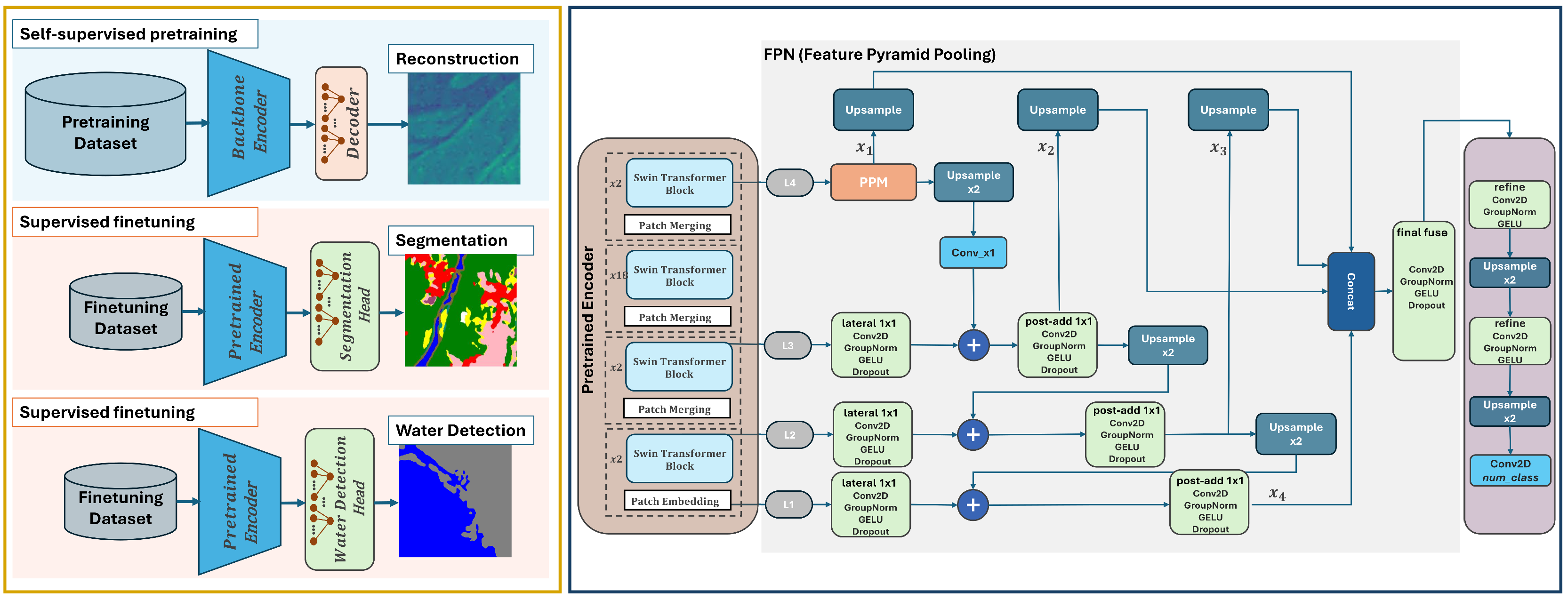}
	\centering
    \caption{Overview of the proposed pipeline. \textit{Left:} self-supervised pretraining on unlabeled ALOS-2 HH using SAR-W-MixMAE \cite{sar_w_mixmae}, followed by supervised finetuning for downstream dense prediction. \textit{Right:} finetuning architecture with a Swin encoder and a UPerNet-style \cite{xiao2018unified} decoder (PPM + FPN), incorporating (i) highest-resolution post-embedding feature injection and (ii) progressive upsampling to improve overall performance while mitigating boundary smoothing.}\label{fig_method}
\end{figure*}

Recently, SAR-focused foundation modeling has begun to emerge. SARCLIP \cite{sarclip2026} explores large-scale vision–language alignment for SAR imagery. SAR-JEPA \cite{sarjepa2024} leverages self-supervised representation learning to capture SAR structure beyond pixel-level reconstruction. SUMMIT \cite{summit2025} combines MIM with auxiliary objectives to better encode spatial layout and scattering-related cues. SARFormer \cite{sarformer_2025_CVPR} further incorporates sensor-specific metadata (e.g., acquisition geometry and imaging modes) to improve robustness to distortions and generalization across diverse SAR collections. These studies indicate that SAR benefits from pretraining strategies tailored to its sensing characteristics. Still, high-quality dense prediction on SAR remains challenging: speckle-like granular patterns can obscure object boundaries, aggressive downsampling can erase thin and slender structures, and long-tailed label distributions commonly found in land use/land cover (LULC) datasets bias learning toward frequent categories. In addition, satellite archives often contain strong spatial and temporal redundancy; naive random sampling can introduce bias in both pretraining and finetuning, reducing sample efficiency and robustness.
This work targets ALOS-2 single-polarization (HH) imagery for Japan-wide LULC mapping and a companion water detection task using reference water masks. Our prior study \cite{imamoglu2025rssj} adapted SAR-W-MixMAE \cite{sar_w_mixmae} pretraining to ALOS-2 and demonstrated clear benefits over training from scratch. However, the decoder design and feature selection still exhibit typical SAR failure modes: (i) over-smoothing and fragmented predictions along object boundaries, (ii) missed thin and slender structures (e.g., narrow rivers, small water bodies, elongated man-made regions), and (iii) degraded performance on rare classes under long-tailed label distributions.
To address these issues with minimal architectural overhead, we introduce three simple enhancements to the baseline \cite{imamoglu2025rssj}. First, we revise the multiscale feature usage to explicitly include the highest-resolution early representation in the decoding pathway. This is motivated by prior observations that early features retain fine spatial detail relevant to edges and boundaries \cite{Yosinski2014NIPS, hariharan2015hypercolumns}, which is especially valuable when SAR cues are weak and boundaries are easily over-smoothed. Second, we modify the segmentation head to a progressive refinement design that alternates lightweight convolutional refinement and step-wise upsampling, encouraging gradual spatial detail recovery rather than abrupt interpolation, thus yielding sharper segmentation contours and improved small-object completeness. Third, building on our earlier class-imbalance analysis \cite{caglayan2025apsar}, we retain the effective focal+dice combination and introduce an $\alpha$-scaling factor to temper class reweighting strength, enabling more stable optimization and stronger gains on rare categories without sacrificing dominant-class performance. The proposed refinements preserve the simplicity and efficiency of a standard Swin \cite{liu2022swin} + UPerNet-style \cite{xiao2018unified} segmentation pipeline, yet improve boundary fidelity, thin-structure delineation, and rare-class consistency in ALOS-2 LULC mapping, while also benefiting water detection for both large water bodies and narrow rivers.

\section{Method}
\label{sec:method}

We build an ALOS-2 SAR foundation model using SAR-W-MixMAE~\cite{sar_w_mixmae} and finetune it for ALOS-2 LULC segmentation (Fig.~\ref{fig_method}). Our design targets three recurring issues in SAR segmentation: (i) multiplicative speckle and spatially varying measurement reliability, (ii) boundary over-smoothing and loss of fine structures, and (iii) class imbalance that hurts rare-class learning.

We pretrain on unlabeled ALOS-2 HH imagery with SAR-W-MixMAE~\cite{sar_w_mixmae}, which adopts the mixed masked autoencoding strategy of MixMAE~\cite{liu2023mixmae}.
MixMAE forms mixed inputs before the encoder and reconstructs the original (unmixed) targets at the decoder via a dual reconstruction objective~\cite{liu2023mixmae}. To better reflect SAR radiometric reliability and reduce sensitivity to speckle-dominated regions, SAR-W-MixMAE applies a pixel-wise backscatter power weighting in the reconstruction loss~\cite{sar_w_mixmae}.

The encoder is a hierarchical Swin Transformer~\cite{liu2022swin} with base configuration
$C=(128,256,512,1024)$ channels,
$H=(4,8,16,32)$ attention heads, and
$B=(2,2,18,2)$ blocks per stage.
For LULC segmentation, we replace the pretraining decoder with a UPerNet-style decoder~\cite{xiao2018unified} (PPM + FPN) that aggregates multi-scale encoder features using an FPN-like top-down pathway~\cite{lin2017feature}, following our baseline in~\cite{imamoglu2025rssj}.

Segmentation boundaries in SAR are prone to over-smoothing due to speckle and downsampling in hierarchical encoders. We apply two lightweight yet effective refinements. First, we explicitly include the highest-resolution post–patch-embedding feature map in the multi-scale decoder, together with early Swin stage features. Second, we employ a progressive upsampling head (staged upsampling with intermediate convolutional refinement) instead of a single large interpolation, enabling boundary refinement at higher spatial resolutions.

For the loss objective, we adopt a combined focal+dice loss \cite{caglayan2025apsar} to address class imbalance and improve segmentation accuracy. To control the strength of weighting, we further scale the focal loss weights by a global scaling factor.

Let $n_k$ be the number of pixels of class $k$ in the training set and
$f_k = \frac{n_k}{\sum_{j=0}^{K-1} n_j}$ the class pixel frequency.
We define subtraction-normalized class weights
\begin{equation}
	w_k = 1 - f_k,
	\qquad
	\alpha_k = \frac{w_k}{\sum_{j=0}^{K-1} w_j}.
	\label{eq:alpha_from_freq}
\end{equation}
To control the strength of reweighting, we scale the weights in the focal term by a single factor $\alpha_{\mathrm{scale}}$.

\paragraph{LULC segmentation loss}
Let $\Omega$ be the set of valid (non-ignored) pixels and $N=|\Omega|$.
For $K$-class segmentation, the model outputs logits $\mathbf{z}_n\in\mathbb{R}^K$ and labels $y_n\in\{0,\dots,K-1\}$.
Define softmax probabilities $p_{n,k}=\mathrm{softmax}(\mathbf{z}_n)_k$ and $p_{t,n}=p_{n,y_n}$.
The focal loss is
\begin{equation}
	\mathcal{L}_{\mathrm{focal}}
	=
	\frac{1}{N}\sum_{n\in\Omega}
	\left(
	-\alpha_{\mathrm{scale}}\alpha_{y_n}\,(1-p_{t,n})^{\gamma}\,\log(p_{t,n} + \epsilon)
	\right),
	\label{eq:focal}
\end{equation}
where $\gamma$ is the focusing parameter~\cite{lin2017focal}.
For dice, with one-hot labels $g_{n,k}=\mathbb{I}[y_n=k]$,
\begin{equation}
	\mathcal{L}_{\mathrm{dice}}
	=
	1-\frac{1}{K}\sum_{k=0}^{K-1}
	\frac{
		2\sum_{n\in\Omega} p_{n,k}\,g_{n,k} + \epsilon
	}{
		\sum_{n\in\Omega} p_{n,k} + \sum_{n\in\Omega} g_{n,k} + \epsilon
	},
	\label{eq:dice_multi}
\end{equation}
with $\epsilon>0$~\cite{milletari2016vnet}.
The total loss is
\begin{equation}
	\mathcal{L}
	=
	\lambda_{\mathrm{focal}}\mathcal{L}_{\mathrm{focal}}
	+
	\lambda_{\mathrm{dice}}\mathcal{L}_{\mathrm{dice}}.
	\label{eq:total_multiclass}
\end{equation}
We use
$\alpha_{\mathrm{scale}}=2.25$, $\gamma=1.1$,
$\lambda_{\mathrm{focal}}=0.57$, and $\lambda_{\mathrm{dice}}=0.32$ in our experiments.

\paragraph{Water detection loss}
For water detection, we optimize a binary objective given by the sum of binary cross-entropy and soft dice loss, both evaluated over valid pixels $\Omega$. We set $\lambda_{\mathrm{dice}}=1.0$ and threshold the predicted probabilities at $\mathrm{pred\_thr}=0.5$ at inference.

\section{Experiments}
\label{sec:experiments}
\subsection{Dataset}
We use a national-scale ALOS-2 single-pol HH SAR dataset over Japan with the JAXA LULC semantic map for self-supervised pretraining and supervised segmentation. To improve coverage and reduce class imbalance, we apply label-guided sampling as in our previous baseline work~\cite{imamoglu2025rssj} inspired by stratified dataset construction for foundation models (e.g.,~\cite{jakubik2023foundation}). Concretely, we use the JAXA LULC labels to drive sampling.
JAXA LULC provides 14 categories over 131 GeoTIFF tiles at 10 m resolution ($12,000 \times 12,000$ pixels each; WGS84 lat/lon) \cite{JAXA_LULC_2023}. We sample ~1,000,000 pixel locations by inverse class-frequency to form lat/lon anchors that upweight rare categories. Using ALOS-2 HH acquisitions from Sep–Nov 2022, we extract non-overlapping $256 \times 256$ SAR patches (1,960,037 total) and retain patches containing $\geq$1 anchor; when labels are required, we extract the aligned $256 \times 256$ LULC label patch.
To prevent temporal leakage, we split by month: Oct–Nov for pretraining and Sep for finetuning/evaluation, yielding 328,238 pretraining patches and 123,842 SAR–label pairs, split into 80,497 / 22,291 / 21,054 for train/val/test. For LULC segmentation, we remap 14 classes to 9 by merging five forest classes into one and merging cropland with paddy field. For water detection, we derive binary labels (water vs. non-water) from the same source.

\subsection{Evaluation and Training Details}
For LULC semantic segmentation, we report mean Intersection-over-Union (mIoU) as the primary metric and mean class accuracy (mAcc) as a complementary measure. We also report per-class IoU and per-class accuracy over the categories to show class-imbalance effects. For water detection (water vs. non-water), we report water-class IoU (IoU\_water) along with precision and recall.

All ALOS-2 inputs are single-channel HH patches of size $256\times256$, normalized using the training-set mean and standard deviation. We pretrain the backbone with SAR-W-MixMAE \cite{sar_w_mixmae} on unlabeled data using AdamW (weight decay $0.05$). The base learning rate is $1.5\times10^{-4}$, scaled linearly by $(global\_batch\_size)/256$, with a cyclic schedule that decays to 0 and 40 warm-up epochs. Unless stated otherwise, pretraining runs for 600 epochs. Pretraining is performed on ABCI 3.0 
with PyTorch DistributedDataParallel (NCCL) on a single node with 8× NVIDIA H200 GPUs (batch size 256 per GPU; global batch size 2048).

For downstream tasks, we finetune the pretrained encoder with a UPerNet-style decoder using AdamW (weight decay $0.05$). We use a base learning rate of $6\times10^{-4}$ with layer-wise learning-rate decay $0.7$, a minimum learning rate of $1\times10^{-6}$, 10 warm-up epochs, and 200 finetuning epochs unless stated otherwise. Finetuning uses the same ABCI single-node 8-GPU setup with batch size 64 per GPU (global batch size 512). For the binary water detection head, we use the same settings but finetune for 100 epochs by default. For ablations without self-supervised initialization, we train the same architectures from scratch for 400 epochs for both LULC segmentation and water detection.

As our downstream objective is dense semantic segmentation (vs. the sparse classification-style evaluation in SAR-W-MixMAE \cite{sar_w_mixmae}), we use longer pretraining and finetuning schedules in our setting.

\subsection{Results and Analyses}
\begin{table}[!h]
	\caption{Ablation study of proposed refinements on the validation set(\%).\yes enabled, \no disabled.}
	\begin{center}
		\setlength{\tabcolsep}{1.6em} 
		\def\arraystretch{1.2}
		\begin{adjustbox}{width=\columnwidth}
			\begin{tabular}{cccccc}
				\hline
				Pretraining 	& High-res feat. 	& Refine-Up & $\alpha$-scale & mIoU 	& mAcc  \\ \hline \hline
				\yes			& \no   		& 	\no  	& \no  		 & 47.45 	& 58.34 \\
				\yes   			& \yes			&	\no		& \no  		 & 48.99 	& 59.92  \\
				\yes   			& \yes			&	\yes	& \no  		 & 49.67 	& 60.93   \\ 
				\no   			& \yes			&	\yes	& \yes  	 & 41.18 	& 50.00  \\
				\yes   			& \yes			&	\yes	& \yes  	 & \textbf{50.26} 	& \textbf{61.12}  \\ \hline
			\end{tabular}
		\end{adjustbox}
		\label{table_ablation_val}
	\end{center}
\end{table}

\begin{figure}[t!]
	\centering
	\includegraphics[width = \columnwidth]{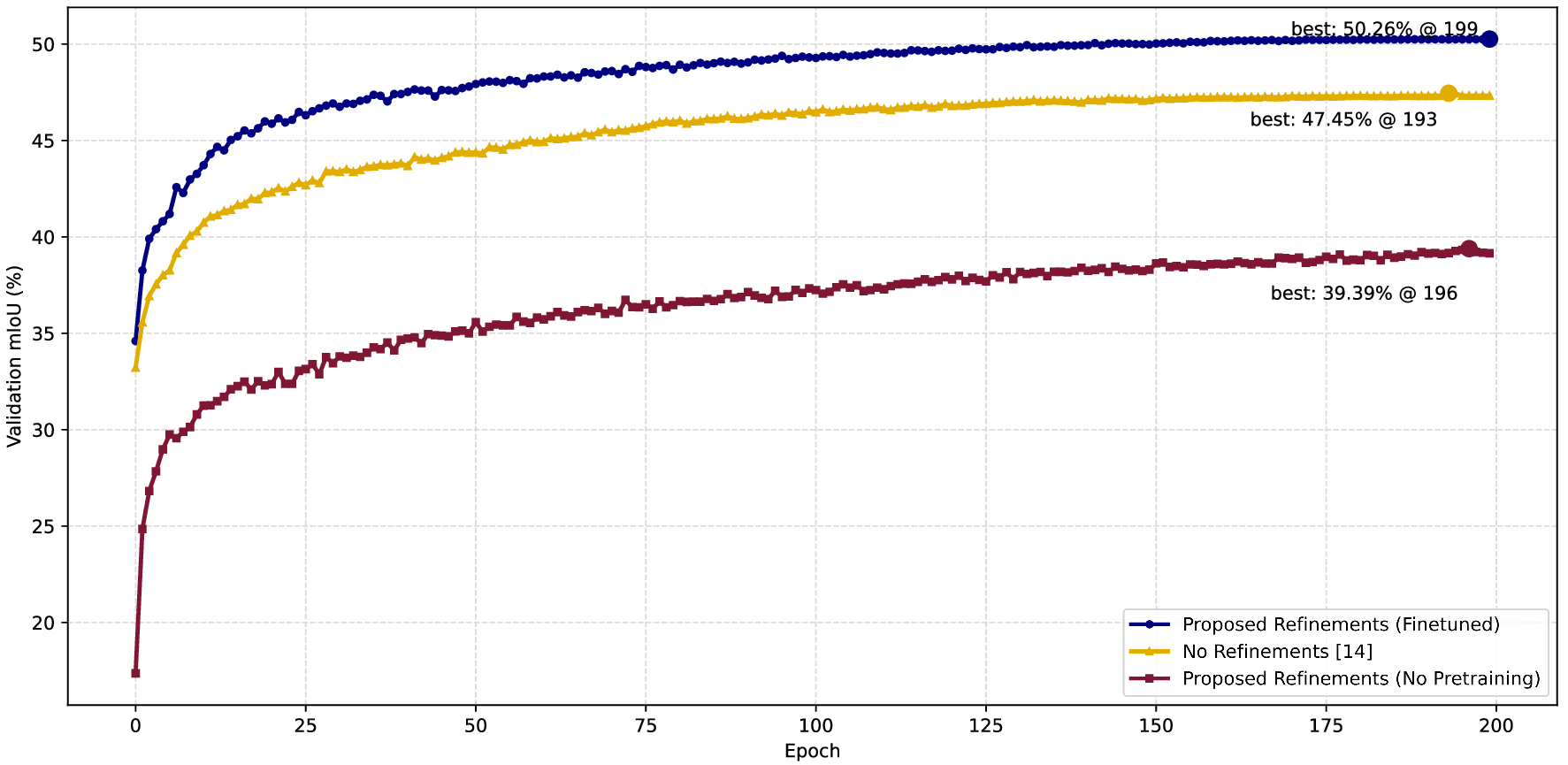}
	\centering
	\caption{Validation mIoU during training; scratch run shown up to 200 epochs.}\label{fig_valmiou}
\end{figure}

\begin{table}[!t]
	\centering
	\caption{Comparison of class-wise IoU and mean IoU on the test set.}
	\renewcommand{\arraystretch}{1.1} 
	\resizebox{\columnwidth}{!}{%
		\begin{tabular}{l|ccccccccc|c}
			\toprule
			Method 		   & Water & Built-up & Cropland & Grassland & Forest & Bare & Solar Panel & Wetland & Greenhouse & mIoU \\
			\midrule
			No Pretraining & 0.89 & 0.67 & 0.56 & 0.29 & 0.85 & 0.19 & 0.03 & 0.13 & 0.08 & 0.41 \\
			RSSJ'25 \cite{imamoglu2025rssj}   & 0.92 & 0.67 & 0.62 & 0.37 & 0.87 & 0.32 & 0.20 & 0.18 & 0.11 & 0.47 \\
			\textbf{This work} 	   & \textbf{0.93} & \textbf{0.70} &\textbf{ 0.65} & \textbf{0.38} & \textbf{0.88} & \textbf{0.33} & \textbf{0.23} & \textbf{0.22} & \textbf{0.17} & \textbf{0.50} \\
			\bottomrule
		\end{tabular}
	}
	\label{tab_iou_results}
\end{table}

\begin{table}[!t]
	\centering
	\caption{Water detection results on the test set(\%).}
	\renewcommand{\arraystretch}{1.2} 
	\resizebox{0.9\columnwidth}{!}{%
		\begin{tabular}{l|ccc}
			\toprule
					   				& IoU\_Water & Precision & Recall \\
			\midrule
			This work (No Pretraining) 	& 90.06 	& 95.60 	& 93.95  \\
			This work (Finetuned)   	& 93.64 	& 96.53 	& 96.91   \\
			\bottomrule
		\end{tabular}
	}
	\label{tab_water_results}
\end{table}

Table~\ref{table_ablation_val} summarizes the validation ablation study. Starting from the pretrained baseline decoder without refinements, we first evaluate the inclusion of high-resolution features after patch embedding, along with adjusted utilization of Swin features in the multi-scale decoder. This yields a clear improvement (+1.54 mIoU, +1.58 mAcc), indicating that injecting higher-resolution features benefits localization. Incorporating Refine-Up (progressive upsampling with intermediate refinement) further improves performance (+0.68 mIoU, +1.01 mAcc), consistent with reduced boundary smoothing. Finally, enabling $\alpha$-scale (tempered class weighting inside the focal term) increases the score to 50.26 mIoU / 61.12 mAcc (+0.59 mIoU, +0.19 mAcc), giving the best overall validation results. Overall, the full set of refinements improves the pretrained baseline by +2.81 mIoU and +2.78 mAcc. The ablation also highlights the impact of self-supervised initialization: without pretraining, even with all refinements enabled, the model reaches 41.18 mIoU / 50.00 mAcc (trained for 400 epochs), which is substantially below its pretrained counterpart.

Fig.~\ref{fig_valmiou} plots validation mIoU over training epochs for three settings: (i) pretrained + proposed refinements, (ii) pretrained without refinements~\cite{imamoglu2025rssj}, and (iii) proposed refinements without pretraining.

The refined pretrained model consistently converges to a higher plateau, achieving a best of 50.26\% (at epoch 199), compared to 47.45\% for the no-refinement baseline. For the no-pretraining run, the curve shown is clipped to the first 200 epochs for a fair visual comparison (our scratch training actually runs for 400 epochs). The corresponding best performance over the full 400-epoch scratch run is reported in Table~\ref{table_ablation_val}.

LULC segmentation (Table~\ref{tab_iou_results}) on the test set shows that this work achieves an mIoU of 0.50, outperforming both no-pretraining variant (0.41, +0.09, +22\% relative) and the prior baseline in \cite{imamoglu2025rssj} (0.47, +0.03, +6.4\% relative). Improvements are observed across categories, with particularly strong gains for under-represented or challenging classes. Relative to the no-pretraining variant, the largest gains are observed for Solar Panel and Bare, followed by consistent improvements in Cropland, Grassland, Wetland, and Greenhouse. Compared to the prior work \cite{imamoglu2025rssj}, gains remain consistent, with the largest margins in Greenhouse and Wetland, indicating improved discrimination of rare classes rather than only boosting frequent classes.
Self-supervised pretraining also benefits binary water detection (Table~\ref{tab_water_results}), even though the pretrained model is trained for only 100 epochs, compared to 400 epochs for the no-pretraining counterpart. Finetuning the pretrained encoder improves IoU\_Water from 90.06 to 93.64, with precision increasing from 95.60 to 96.53 and recall from 93.95 to 96.91. The larger gain in recall suggests pretraining primarily reduces missed-water errors (false negatives), while maintaining high precision.

\begin{figure}[t!]
	\centering
	\includegraphics[width = \columnwidth]{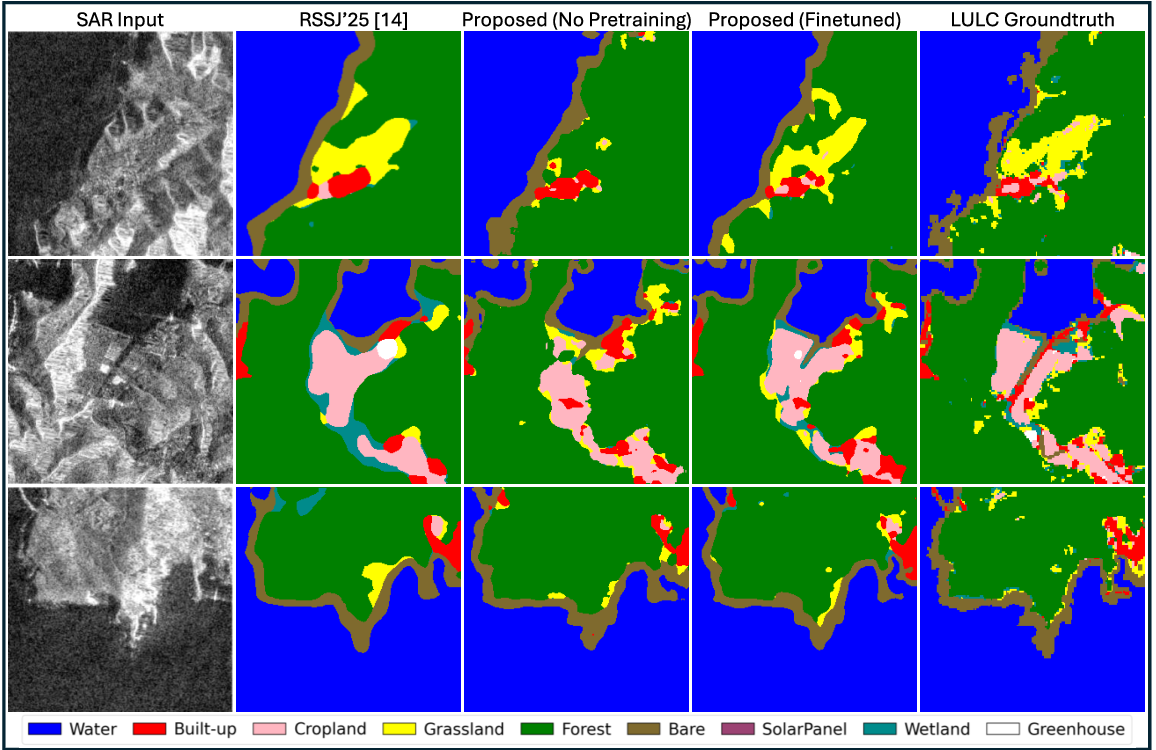}
	\centering
	\caption{Qualitative LULC segmentation comparison among RSSJ'25~\cite{imamoglu2025rssj}, our no-pretraining model, and our pretrained model with proposed refinements.}\label{fig_lulc_results}
\end{figure}
Fig.~\ref{fig_lulc_results} provides qualitative comparisons between RSSJ'25~\cite{imamoglu2025rssj}, our model trained from scratch, and our pretrained+finetuned model. Across all examples, the proposed approach produces sharper boundaries and better preservation of fine structures, reducing the boundary over-smoothing visible in~\cite{imamoglu2025rssj}. In particular, thin and elongated regions (e.g., narrow built-up strips and small land-cover fragments near coastlines) are more consistently recovered, and small-scale details are less likely to be merged into surrounding dominant classes. The improvement is strongest with pretraining, where the predictions more closely match the ground truth in both object extent and fine-grained spatial layout, indicating better localization and rare/ambiguous structure handling.

\begin{figure}[t!]
	\centering
	\includegraphics[width = \columnwidth]{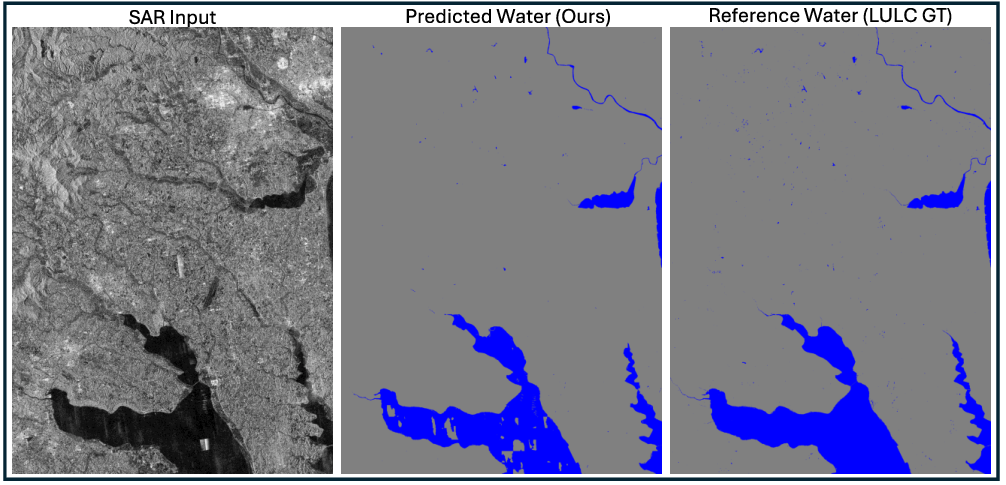}
	\centering
	\caption{Large-area water detection example showing SAR input, model prediction, and reference water mask.}\label{fig_water_results}
\end{figure}
Fig.~\ref{fig_water_results} shows a large-area qualitative example for water detection. The model captures both large contiguous water bodies (coast/sea) and fine-scale structures such as thin river streams and small inland water patches, closely matching the reference mask. Most discrepancies occur near the coastline and in scattered small regions, where the SAR backscatter suggests ambiguous boundaries; these errors are consistent with likely label–signal mismatch (e.g., shoreline ambiguity, temporal differences, or annotation noise) rather than systematic false detections.

\section{Conclusion}
\label{sec:conc}
We presented simple model refinements for dense LULC segmentation on ALOS-2 HH SAR, targeting boundary fidelity and rare-class robustness. Incorporating high-resolution features, a progressive refine-up head, and an $\alpha$-scaled focal+dice objective yields consistent gains on the Japan-wide benchmark and improves water detection performance. Future work will focus on better separating man-made structure categories and mitigating limitations of single-polarization SAR.


\small
\bibliographystyle{IEEEtranN}
\bibliography{egbib}

\end{document}